\documentclass[sigconf]{acmart}

\AtBeginDocument{%
  \providecommand\BibTeX{{%
    \normalfont B\kern-0.5em{\scshape i\kern-0.25em b}\kern-0.8em\TeX}}}

\copyrightyear{2022}
\acmYear{2022}
\setcopyright{rightsretained}
\acmConference[AHs 2022]{Augmented Humans 2022}{March 13--15,
2022}{Kashiwa, Chiba, Japan}
\acmBooktitle{Augmented Humans 2022 (AHs 2022), March 13--15,
2022, Kashiwa, Chiba, Japan}\acmDOI{10.1145/3519391.3524034}
\acmISBN{978-1-4503-9632-5/22/03}

\acmPrice{15.00}
\acmISBN{978-1-4503-XXXX-X/18/06}

\usepackage{enumitem}
\usepackage{amsmath} 
\usepackage{multirow}
\usepackage[acronym,shortcuts]{glossaries}
\newacronym{CNN}{CNN}{convolutional neural network}
\newacronym{FNN}{FNN}{feedforward neural network}
\newacronym{PSE}{PSE}{perceptual stimulus encoder}
\newacronym{SPV}{SPV}{simulated prosthetic vision}
\newacronym{VPU}{VPU}{video processing unit}

\begin{document}

\title{Deep Learning--Based Perceptual Stimulus Encoder for Bionic Vision}


\author{Lucas Relic}
\affiliation{
    \institution{University of California, Santa Barbara}
    \city{Santa Barbara}
    \state{CA}
    \country{USA}
}
\email{lucasrelic@ucsb.edu}

\author{Bowen Zhang}
\affiliation{
    \institution{University of California, Santa Barbara}
    \city{Santa Barbara}
    \state{CA}
    \country{USA}
}
\email{bowen68@ucsb.edu}

\author{Yi-Lin Tuan}
\affiliation{
    \institution{University of California, Santa Barbara}
    \city{Santa Barbara}
    \state{CA}
    \country{USA}
}
\email{ytuan@ucsb.edu}

\author{Michael Beyeler}
\affiliation{
    \institution{University of California, Santa Barbara}
    \city{Santa Barbara}
    \state{CA}
    \country{USA}
}
\email{mbeyeler@ucsb.edu}

\renewcommand{\shortauthors}{Relic et al.}

\begin{abstract}
Retinal implants have the potential to treat incurable blindness, yet the quality of the artificial vision they produce is still rudimentary.
An outstanding challenge is identifying electrode activation patterns that lead to intelligible visual percepts (phosphenes).
Here we propose a \acf{PSE} based on \acfp{CNN} that is trained in an end-to-end fashion to predict the electrode activation patterns required to produce a desired visual percept.
We demonstrate the effectiveness of the encoder on MNIST using a psychophysically validated phosphene model tailored to individual retinal implant users.
The present work constitutes an essential first step towards improving the quality of the artificial vision provided by retinal implants. 
\end{abstract}



\begin{teaserfigure}
    \centering
    \includegraphics[width=0.8\columnwidth]{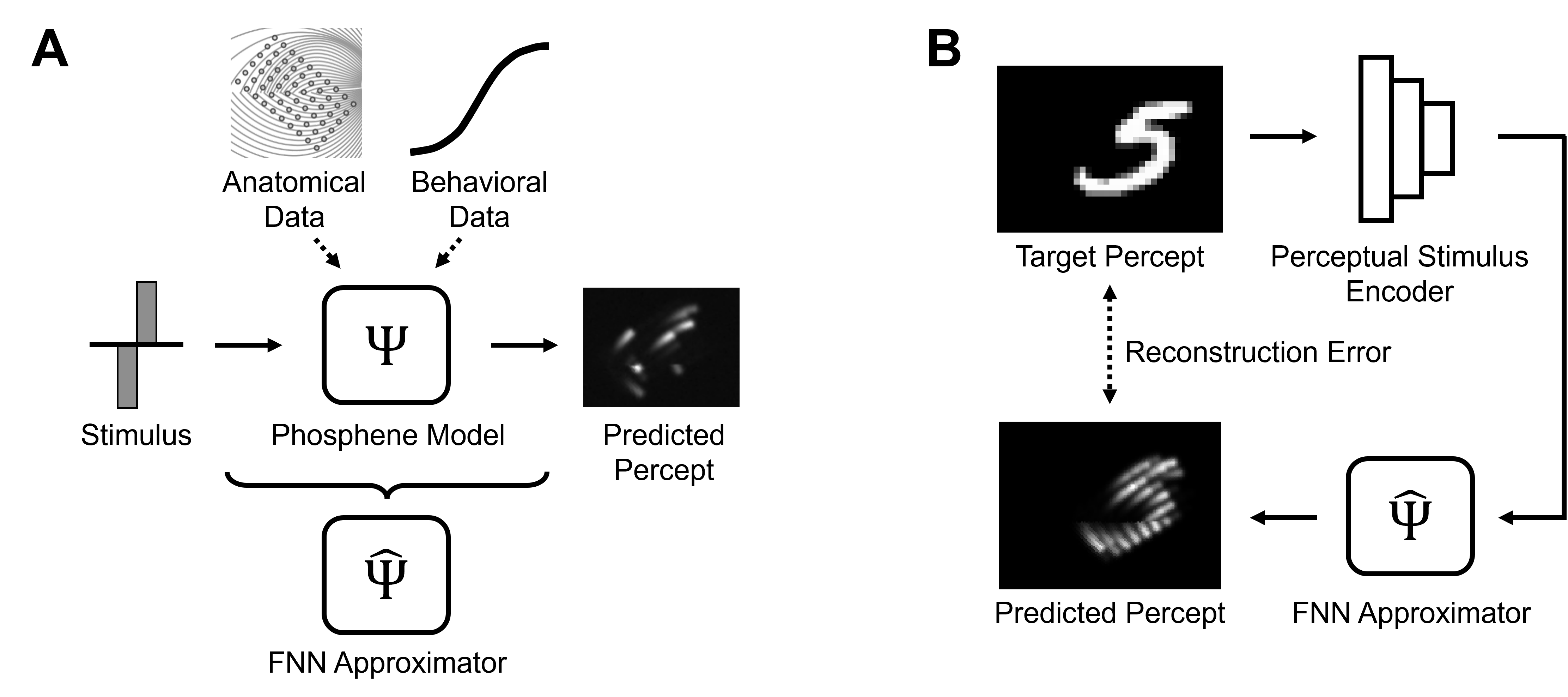}
    \caption{
        A) \Acf{SPV}. Constrained by neuroanatomical and/or psychophysical data, a phosphene model $\Psi$ (e.g., \cite{beyeler2019axon}) predicts what a retinal implant user should ``see'' for any given input stimulus.
        The predicted percept is typically a nonlinear continuous function of the input stimulus, thus $\Psi$ can be approximated by a generic \acf{FNN}, $\hat{\Psi}$, which is amenable to differentiation.
        B) End-to-end optimization of bionic vision.
        For a given target percept, a stimulus encoder based on a \acf{CNN} is trained to predict the combination of active electrodes that generates the percept with the smallest possible reconstruction loss.
        The \acs{FNN} approximator is fixed during encoder training.
    }
    \label{fig:overview}
\end{teaserfigure}

\maketitle

\section{Introduction}
Hereditary retinal diseases such as retinitis pigmentosa 
are among the leading causes of incurable blindness in the world.
Retinal implants 
(though elementary) provide an improved ability to localize high-contrast objects as well as perform basic orientation \& mobility tasks~\cite{ayton_update_2020}.
These devices electrically stimulate surviving cells in the visual pathway to evoke visual percepts (phosphenes). 

However, the quality of this artificial vision is still rudimentary, as the visual percepts elicited by current implants are often unrecognizable~\cite{erickson-davis_what_2021}.
A major outstanding challenge is identifying electrode activation patterns that lead to perceptually intelligible phosphenes.
One approach is to consider this an end-to-end optimization problem, where a deep neural network (encoder) is trained to predict the electrical stimulus needed to produce a desired percept (Fig.~\ref{fig:overview}).

To this end, we make the following contributions:
\begin{itemize}[topsep=0pt]
    \item We propose a \acf{PSE} based on \acfp{CNN} that is trained in an end-to-end fashion to predict the electrode activation patterns required to produce a desired visual percept. Importantly, the encoder is based on 
    a psychophysically validated computational model of bionic vision \cite{beyeler2019axon}.
    \item We demonstrate the effectiveness of the \acs{PSE} on the MNIST dataset for three different users of the Argus II Retinal Prosthesis System (Second Sight Medical Products) \cite{luo_argusr_2016}.
   
\end{itemize}


\section{Related Work}
\label{sec:related_work}

\subsection{Simulated Prosthetic Vision (SPV)}
The goal of \ac{SPV} is to predict what bionic eye users ``see'' in response to electrical stimulation.
To date, most \ac{SPV} studies rely on the scoreboard method, which assumes that each phosphene acts as a small independent light source, analogous to the images projected on the light bulb arrays of some sports stadium scoreboards \cite{dobelle_artificial_2000}.
However, evidence suggests that phosphenes often appear distorted (e.g., as simple geometric shapes such as lines, wedges, and blobs) and vary drastically across subjects and electrodes \cite{erickson-davis_what_2021}.
More recently, \ac{SPV} models have therefore aimed to explain phosphene appearance as a function of neuroanatomical and psychophysical data \cite{beyeler2019axon,granley_computational_2021} (Fig.~\ref{fig:overview}A).
Open-source implementations for many of these models are provided by \emph{pulse2percept}, a Python-based bionic vision simulator \cite{michaelbeyeler2017}.

\subsection{End-to-End Optimization of Bionic Vision}
Although deep learning has previously been combined with \ac{SPV} to perform image processing on the predicted percept (e.g., \cite{8695610,horne2016semantic,han_deep_2021}), only few studies have considered the quest to improve the quality of artificial vision as an end-to-end optimization problem (Fig.~\ref{fig:overview}B).
Most notably, van Steveninck et al.~\cite{van2020end} used a similar approach to ours by training an encoder-decoder deep neural network with a scoreboard model in the loop.
However, their approach differs in three crucial ways: 
i) their phosphene model could not account for empirical data from current retinal implant users \cite{beyeler2019axon,erickson-davis_what_2021},
ii) their loss function did not consider the performance of either encoder or phosphene model, and most importantly
iii) their decoder was in itself a deep neural network that could potentially learn to compensate for any deficiencies in the encoder or the phosphene model.


\section{Methods}
\label{sec:methods}

Our model is illustrated in Fig.~\ref{fig:overview} and described in more detail below.
We first approximated a psychophysically validated phosphene model ($\Psi$, \cite{beyeler2019axon}) with a generic \ac{FNN} ($\hat{\Psi}$, Fig.~\ref{fig:overview}A).
Once trained, the weights of the \ac{FNN} Approximator were frozen and used to train a \ac{PSE} to minimize the reconstruction error between predicted and target percepts (Fig.~\ref{fig:overview}B).
The \ac{PSE} took a target image as input and returned a combination of active electrodes as output, which were then fed into the \ac{FNN} Approximator to predict a visual percept.
The pixel-wise mean squared error between predicted and target percept served as the reconstruction error.
The error was then backpropagated via the differentiable FNN Approximator to update the weights of the \ac{PSE}.
As a proof of concept, we considered the model's ability to predict handwritten digits from the popular MNIST dataset.

\subsection{FNN Approximator}
Our group~\cite{beyeler2019axon} demonstrated through computational modeling that phosphene shape in epiretinal implants primarily depends not just on stimulus parameters but also on the retinal location of the stimulating electrode.
This model ($\Psi$) can be fit to individual patients; however, it is not differentiable.
We therefore approximated $\Psi$ with a single-layer \ac{FNN} ($\hat{\Psi}$) that took a $1 \times 60$ vector of current amplitudes as input (one amplitude for each electrode in the Argus II Retinal Prosthesis System) and returned a $121 \times 161$ image as output (i.e., the predicted percept).
The FNN Approximator was trained on a synthetic dataset (50,000 samples, 80-20 train-test split) generated with $\Psi$: each sample was generated by first randomly selecting a number $N \in [1, 30]$ to stimulate, then randomly assigning a stimulation current between 1 and 10 microamps to each electrode.
After training, the weights of the \ac{FNN} Approximator were frozen.

\subsection{Perceptual Stimulus Encoder (PSE)}
The \ac{PSE} was a \ac{CNN} consisting of two $3 \times 3$ convolutional layers (stride 1) followed by a max pooling layer after each convolutional layer, and a fully connected layer at the end.

Rather than optimizing the \ac{PSE} with a pre-trained \ac{FNN} Approximator in the loop, one might also consider training an inverse phosphene model ($\hat{\Psi^{-1}}$) to directly predict the required stimulus for a desired percept.
We therefore trained an inverse model of equal depth to the \ac{PSE} as a baseline model.

\subsection{Simulated Bionic Eye Users}
Since phosphene appearance varies drastically across patients, we followed Beyeler et al.~\cite{beyeler2019axon} to tailor the phosphene model $\Psi$ to the individual implant setup of three Argus II patients: 12-005, 51-009, and 52-001.
The setup for these patients mainly differed in the implant location and in the values of model parameters $\rho$ (describing phosphene size) and $\lambda$ (describing phosphene elongation) \cite{beyeler2019axon}.
Consequently, we had to train three different FNN Approximators and three different \acp{PSE}.

\begin{table*}[!t]
\begin{tabular}{@{}llll@{}}
\toprule
             & Subject 12-005 & Subject 51-009 & Subject 52-001 \\ \cmidrule(l){2-2} \cmidrule(l){3-3} \cmidrule(l){4-4}
Perceptual Stimulus Encoder (PSE)  & $\mathbf{0.0317 \pm 0.014}$         & $\mathbf{0.0547 \pm 0.019}$        & $\mathbf{0.0311 \pm 0.012}$       \\
Inverse Phosphene Model ($\hat{\Psi}^{-1}$) & $0.0371 \pm 0.016$    & $0.0718 \pm 0.025$    & $0.0456 \pm 0.017$    \\ \bottomrule
\end{tabular}
\caption{Reconstruction error (pixel-wise mean squared error) achieved on MNIST, reported as mean $\pm$ standard deviation across samples in the test set.}
\label{tab:results}
\end{table*}

\begin{figure*}[t]
    \centering
    \includegraphics[width=0.8\textwidth]{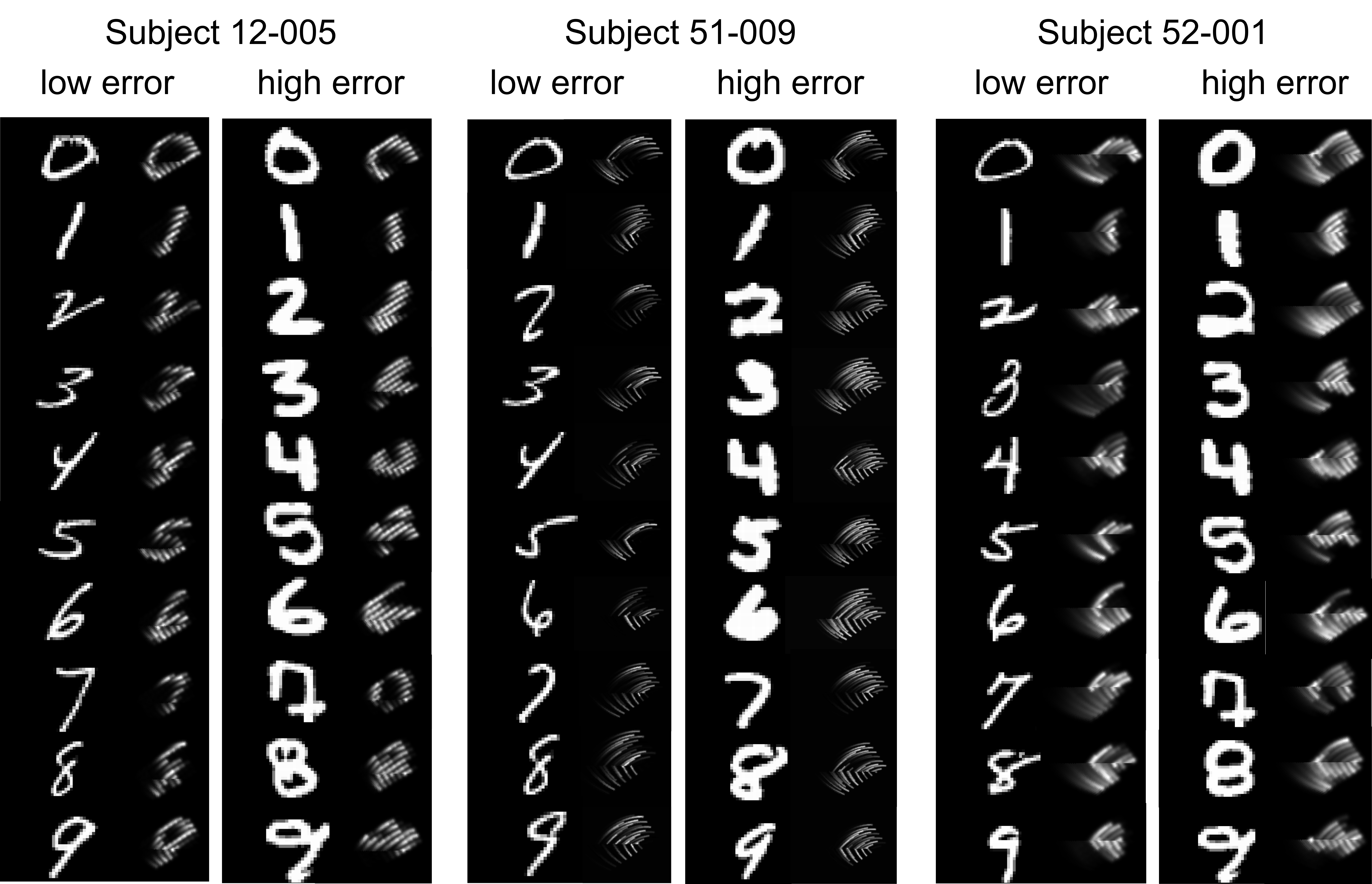}
    \caption{Representative example predictions achieved by the \ac{PSE} trained on three different phosphene models that represent three different Argus II subjects. For each digit (left column in each panel), the corresponding predicted percept (right column) is shown.}
    \label{fig:results}
\end{figure*}

\section{Results}
\label{sec:results}

Results are summarized in Table~\ref{tab:results}.
The best reconstruction error was achieved for Subject 52-001 ($0.0311)$, followed by 12-005 ($0.0317$) and 51-009 ($0.0547$).
The relatively poor performance of the model for Subject 51-009 may be due to the fact that this subject sees very thin, elongated phosphenes that may not be easily combined to form an MNIST digit.
A deeper encoding model may help to overcome this issue.

The \ac{PSE} outperformed the inverse phosphene model in all three instances.
While and end-to-end trained inverse phosphene model might seem advantageous at first, it has the notable drawback that the mapping from desired percept to required stimulus may be one-to-many, which may not lend itself well to gradient-based optimization.


Fig.~\ref{fig:results} shows representative examples of digits with low (good) and high (bad) reconstruction errors for all three subjects. The PSE was able to utilize the inherent streakiness of the phosphene model to produce recognizable digits with a small number of active electrodes. 
However, it is worth noting that the digits with the lowest reconstruction error did not always look the best.
The model tended to yield poor results for thick digits, which required the activation of a large number of electrodes and subsequently produced large indistinct blobs.
Since all phosphenes have an inherent orientation, the model also struggled with digits whose edges ran orthogonal to that orientation.



\section{Conclusion}
The present work constitutes an essential first step towards improving the quality of artificial vision provided by current retinal implants.
Future work should focus on more naturalistic datasets and developing 
a better perceptual error metric.
The most important future contribution, however, will be to demonstrate that the proposed approach is a viable strategy to improve the quality of artificial vision in real bionic eye users.

\begin{acks}
This work was supported by the National Institutes of Health (NIH R00 EY-029329 to MB).
\end{acks}

\newpage
\bibliographystyle{ACM-Reference-Format}
\bibliography{references.bib}


\end{document}